# De-Noising and Segmentation of Epigraphical Scripts

P Preeti

*Assistant Professor*
*Department of Computer Science*
*PES University*
*Bangalore*
preethip@pes.edu

Hrishikesh Viswanath

*Department of Computer Science*
*PES University*
*Bangalore*
hrishikeshv@pesu.pes.edu

*Abstract*—This paper is a presentation of a new method for denoising images using Haralick features and further segmenting the characters using artificial neural networks.

The image is divided into kernels, each of which is converted to a GLCM (Gray Level Co-Occurrence Matrix) on which a
Haralick Feature generation function is called, the result of which is an array with fourteen elements corresponding to fourteen features
The Haralick values and the corresponding noise/text classification form a dictionary, which is then used to de-noise the image through kernel comparison.

Segmentation is the process of extracting characters from a
document and can be used when letters are separated by white space, which is an explicit boundary marker. Segmentation is the first step in many Natural Language Processing problems. This paper explores the process of segmentation using Neural Networks.

While there have been numerous methods to segment characters of a document, this paper is only concerned with the accuracy of doing so using neural networks.

It is imperative that the characters be segmented correctly, for failing to do so will lead to incorrect recognition by Natural language processing tools.

Artificial Neural Networks was used to attain accuracy of upto 89%. This method is suitable for languages where the characters are delimited by white space. However, this method will fail to provide acceptable results when the language heavily uses connected letters. An example would be the Devanagari script, which is predominantly used in northern India.

## I. INTRODUCTION

### A. Denoising

Epigraphs is a term to denote inscriptions, usually on statues, buildings or coins. This paper focuses on segmentation of Epigraphical scripts, specifically epigraphs depicting Brahmi scripts of 3rd and 4th century BCE.

The images of Epigraphical scripts are inherently noisy. This noise cannot be arbitrarily classified into any class of noises. Nothing about the noise is known, except that it exists in the image. It is quite hard to decipher the text unless the noise is removed. Decrypting scripts are an essential component in unearthing History, and this is not possible unless the images are legible. De-Noising plays an important role in understanding these Epigraphical scripts.

A Brute Force approach is employed in detection, classification and the subsequent removal of noise. It is important to ensure that no part of the text is de-noised for it will increase the illegibility of the document. Haralick Features or specifically, the Maximum Correlation Coefficient [1] is to identify noise.

### B. Segmentation

The proposed method uses digital images of Epigraphical scripts as the input for segmentation. Epigraphical scripts comprise text inscribed on buildings, statues or coins. The text is not linearly arranged as in case of printed documents, but can be skewed. Letter spacing is not consistant and neither is the dimensions of the characters.

The script uses diacritical marks to differentiate between vowels and consonants. Each character is delimited by white space, which acts as border for black characters.

Segmentation is the process of extracting letters from a stream of characters. The first step of segmentation is data cleaning, which removal of noise from the document.

The complexity arises when there are overlapping characters. In such cases, the neural network cannot incorrectly segment the characters. Doing so will lead to problems with character recognition.

Epigraphical text is harder to segment into letters due to varying and uneven character dimensions. To remedy this issue, the text is first segmented using regular overlapping rectangular kernels. The dimension of the rectangle is chosen in such a way that it encompasses one character.

## II. FEATURES OF BRAHMI SCRIPT

Brahmi script is said to have originated in India in 3rd or 4th Century BCE. The script is widely used even today in southern India. Scripts of modern South Indian languages like Kannada or Telugu have formats that closely resemble Brahmi script. This method can be applied on any script that mimics Brahmi. In Brahmi script, Words are usually written from left to right. However, There have been instances of the text running right to left similar to Aramic languages.

Each letter of the Brahmi script represents a consonant. Vowel sounds are depicted by appending symbols to the consonant. It is assumed that a vowel follows every consonant.

Fig. 1. Consonants of Brahmi Script

Fig. 2. Vowels and their usage

Fig. 3. characters denoting numbers

## III. HARALICK FEATURE EXTRACTION

It is impossible to differentiate between noise and text if the kernel contains both text and noisy areas. Choosing the size of the kernel such that it is big enough to span the regions of noise but not so big that it encompasses both text and noise is the basic step of the process. In this case, the kernel size was set to 20x20. It is to be noted that Kernel does not refer to the null space of the image, but merely a sliding window that slides through the image such that at any given instant, the kernel contains a certain 20x20 sub matrix of the image.

The RGB Kernel is converted to a Grey Level Co-Occurrence Matrix [2]. The GLCM refers to the distribution of co-occurring gray scale pixel values. GLCM is a measure of grey scale distribution of the image. They are sensitive to local interaction between the pixels. GLCMs are used to study the texture of the image through spatial interactions between the pixels.

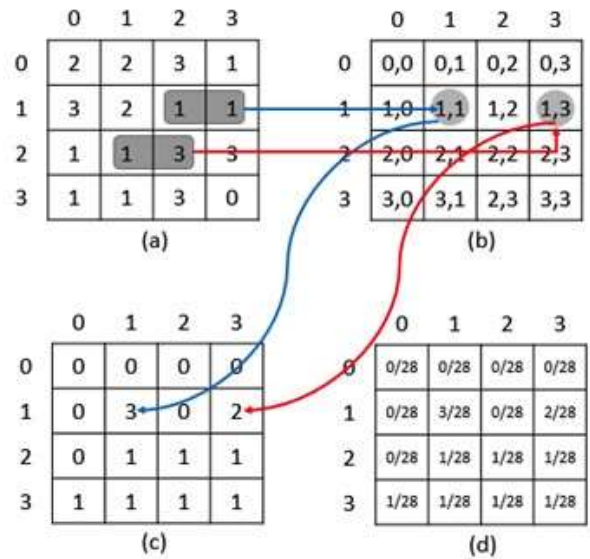

Fig. 4. Generating GLCM from an RGB Matrix

$$\bar{Q}(i,j) = \sum_k \frac{p(i,k)p(j,k)}{p_x(i)p_y(k)}$$

Fig. 5. Formula to calculate Q from GLCM

An Iterative Approach is followed in generating the Haralick Features [3] from the GLCM. For each pixel, four directions are specified, with respect to its adjacent pixels – horizontal, vertical and the left – right diagonals.

Haralick features, as specified by Robert Haralick are Angular Second Moment, Contrast, Correlation, Variance, Inverse Difference Moment, Sum Average, Sum Variance, Sum Entropy, Entropy, Difference Variance, Difference Entropy, Measure of Correlation I II and Maximum Correlation Coefficient.

To denoise the image, we have considered Maximum Correlation Co-efficient as the reference for noise profiling.

Maximum Correlation Coefficient refers to the square root of the second largest value of Matrix Q, where Q is calculated from the GLCM Matrix p using the given formula.

### IV. Building the dictionary and De-Noising

The Maximum Correlation Co-efficient values corresponding to the GLCMs of about 200 kernels of the image are written into an excel document.

| Value | Classification |
|---|---|
| 0.88917 | Text |
| 0.88876 | Text |
| 0.8885 | Text |
| 0.88832 | Mostly Text |
| 0.88861 | Mostly Text |
| 0.88845 | Mostly Text |

Fig. 6. Dictionary of Haralick feature values and the corresponding classification

Using the kernels, these values are manually classified into noise, Mostly Noise, Mostly Text and Text. A Scatter plot of values corresponding to noise, text etc. is plotted. This gives a range of values for each of these categories. The process of generating Haralick features on the GLCM is repeated on the images. Only this time, the values are compared with the range corresponding to noise. If the noise is identified, it is replaced by a black box of dimensions 10x10. This method while not very efficient, succeeds in removing Noise that is structurally (Radius, thickness) different from text but fails to remove noise which is similar to the text.

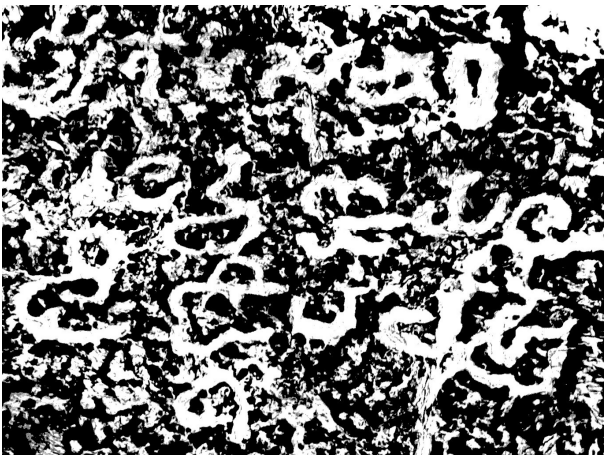

Fig. 7. Noisy Image

### V. Segmentation

The method proposed in this paper is the Recognition based Approach to segmentation [3]. Recognition based approach refers to the method where the system looks for components that can be used to segregate the image into classes. Many Segmentation methods suggested so far rely on heuristic rules to classify the segments. In this method however, there is no heuristic to split the set of samples into classes.

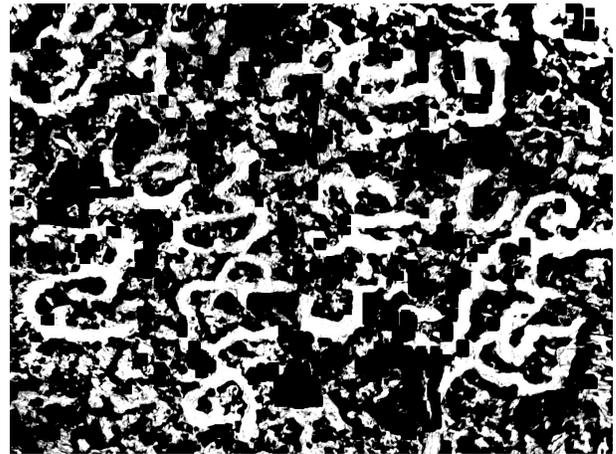

Fig. 8. Image with reduced noise

#### A. Pre-processing

The neural network takes as input, the data set of rectangular segments encompassing fully segmented or partially segmented characters. The image is first segmented using a regular sized rectangular kernel such that it fits a single character. This kernel is slid across the image and captures a region of the image in each iteration.

This process continues until the kernel reaches the end of the image. Once this is done, the next step is to split the input into two classes - Valid characters and invalid characters. Using the predefined Brahmi character set, the segments are manually put into one of these two classes. This forms the input for the neural network.

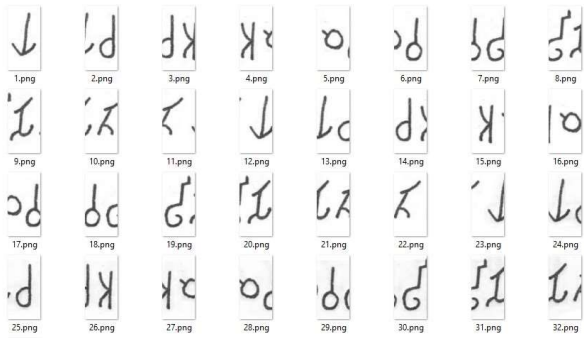

Fig. 9. Input data set comprising regular sized segments

### B. Artificial Neural Network

A feed forward neural network is used for segmentation. The network uses ReLU activation function for all of its hidden neurons. The reason for doing so is that ReLU does not activate all the neurons at once. The sparse activation of neurons makes the network efficient and easy for computing.

For any input x, the ReLU function can have two output values - 0 and x. If the input is either zero or negative, the output is 0. Otherwise, it is x.

The network has one input layer, followed by one or more hidden layer with ReLU neurons. Finally, there is one output layer. Multiple layers aid in learning linear and non linear relationships between input and output [4].

Here is a diagram depicting a feed forward neural network.

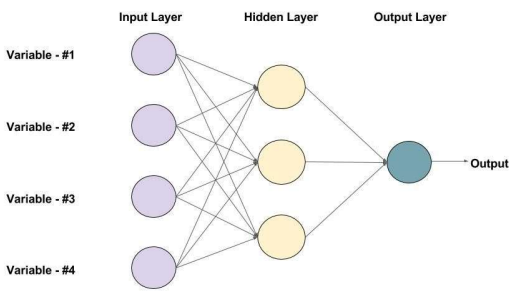

Fig. 10. Feed Forward Neural Network

The network "Learns" to categorize segments through training [5]. By adjusting weights and the number of iterations, the accuracy of classifying can be improved. Care is taken to ensure that over fitting is minimized.

Over fitting is a type of error that occurs when a model manages to fit the data points too closely, achieving high levels of accuracy for that instance of data but ultimately fails to accurately classify other data points not in the original set.

In this scenario, the output is known beforehand, which makes it appropriate to use supervised learning to train the neural network. This process involves comparison of the output generated by the neural network against the expected output. Furthermore, back propagation, a gradient based optimization technique is used to adjust the weights with each iteration. Back propagation aids with minimizing error in predicting the output.

The process of classifying is split into training and testing. The set of segments is first manually classified into valid and non valid characters using the database of valid characters. Upon doing so, the segments are further split such that 80 percent of the segments are used for training, while the remaining 20 percent of it is used for testing.

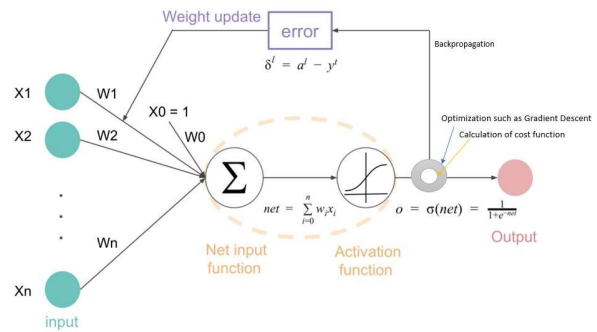

Fig. 11. Back Propagation

Both the class and the images are provided as input to the neural network. The weight is initially set to 0.90. The network iterates 2000 times and classifies the segments with an accuracy of 89%.

The advantage of using neural network is that it doesn't require any heuristic to classify the images. Manual classification, however is the only prerequisite for this method. Manual classification requires that the user know the language beforehand.

This method works in all cases where the letters are disjoint. Languages which heavily make use of connected letters cannot be segmented using this method.

If an image was classified as a valid character, then that image was explicitly written to disk. The final set of segmented characters was written to an output folder.

## VI. OUTPUT

The output obtained is shown in figures 7 and 8. The first image consists of the set of correctly labeled characters. The second image is a graph of cost against iteration count. Also shown in the figure is the accuracy of the model for training data. The accuracy is found to be 89%.

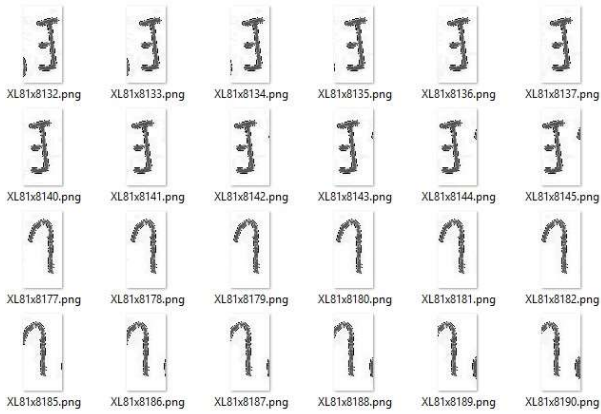

Fig. 12. Set of segments labelled as characters

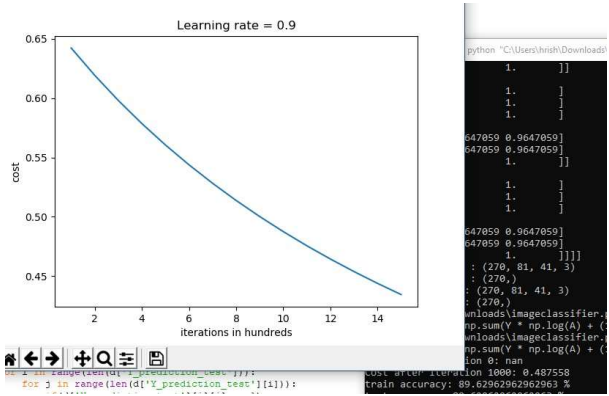

Fig. 13. Graph of cost vs. iterations when the weight is 0.9

VII. CONCLUSION

Haralick Features are effective in denoising images where the noise is sparse and has a different texture from the text but fails to differentiate between noise and text when the noise has the same texture as text.

Artificial Neural Networks can be used to segment the characters of an Epigraph with an accuracy of 89%.

This method is effective for languages with white space delimited characters.